%% file: main.tex
\documentclass[letterpaper, 10 pt, journal, twoside]{IEEEtran}
\IEEEoverridecommandlockouts
\usepackage[ruled,vlined,linesnumbered]{algorithm2e}
\usepackage{cite}
\usepackage{amsmath,amssymb,amsfonts,algorithmic,textcomp, multirow}
\usepackage{graphicx,subfig,float,xcolor,url,hyperref,etoolbox}
\usepackage{tikz}
\usetikzlibrary{positioning}
\usepackage[nolist]{acronym}
\hypersetup{
    colorlinks,
    linkcolor={blue!50!black},
    citecolor={blue!50!black},
    urlcolor={black!80!black}
}

\usepackage{tabularx}
\usepackage{soul}
\usepackage{adjustbox}
\usepackage{svg}
\usepackage{array}
\usepackage{multirow}

\def\BibTeX{{\rm B\kern-.05em{\sc i\kern-.025em b}\kern-.08em
    T\kern-.1667em\lower.7ex\hbox{E}\kern-.125emX}}
    
\makeatletter
  \patchcmd{\@maketitle}
   {\addvspace{0.5\baselineskip}\egroup}
   {\addvspace{-0.1\baselineskip}\egroup}
   {}
   {}
\makeatother
\SetKw{KwBy}{by}
\SetKw{KwNot}{not}
\makeatletter
\newcommand\notsotiny{\@setfontsize\notsotiny\@vipt\@viipt}
\makeatother

\begin{document}

    % \title{\LARGE Underwater 6D Object Pose estimation and Dataset collection for Manipulation}
    \title{\LARGE Model-Based Underwater 6D Pose Estimation from RGB}
    
    \author{Davide Sapienza$^{\dag}$, Elena Govi$^{\dag}$, Sara Aldhaheri$^{*}$, Marko Bertogna$^{\ddag \dag}$, Eloy Roura$^{*}$,\\ {\`E}ric Pairet$^{*}$, Micaela Verucchi$^{\ddag}$, Paola Ard{\'o}n$^{*}$ 
    \vspace{-0.5cm}
    \thanks{Manuscript received: February 10, 2023; Revised May 16 and August 8, 2023; Accepted September 7, 2021. This paper was recommended for publication by Editor Tamim Asfour upon evaluation of the Associate Editor and Reviewers' comments.}
    \thanks{$^{*}$TII ARRC. Abu Dhabi, UAE. $\ddag$Hipert s.r.l and $^{\dag}$University of Modena and Reggio Emilia. Modena, Italy.
    {\tt davide.sapienza@unimore.it}. Authors thank SpinItalia for supporting the experimental campaign in Italy.}
    }
    
    \input{acronyms}
    \markboth{IEEE Robotics and Automation Letters. Preprint Version. Accepted September, 2023}{Sapienza \MakeLowercase{\textit{et al.}}: Model-Based Underwater 6D Pose Estimation from RGB}
    \maketitle
    
    \begin{abstract}
        Object pose estimation underwater allows an autonomous system to perform tracking and intervention tasks. Nonetheless, underwater target pose estimation is remarkably challenging due to, among many factors, limited visibility, light scattering, cluttered environments, and constantly varying water conditions. An approach is to employ sonar or laser sensing to acquire 3D data, however, the data is not clear and the sensors expensive.
        For this reason, the community has focused on extracting pose estimates from RGB input.
        In this work, we propose an approach that leverages 2D object detection to reliably compute 6D pose estimates in different underwater scenarios. 
        We test our proposal with $4$ objects with symmetrical shapes and poor texture spanning across $33{,}920$ synthetic and $10$ real scenes. All objects and scenes are made available in an open-source dataset that includes annotations for object detection and pose estimation.
        When benchmarking against similar end-to-end methodologies for 6D object pose estimation, our pipeline provides estimates that are $\sim\!\!\!8\%$ more accurate.
        We also demonstrate the real-world usability of our pose estimation pipeline on an underwater robotic manipulator in a reaching task.
    \end{abstract}

    
    %%%%%%%%%%%%%%%%%%%%%%%%%%%%%%%%%%%%%%%%%%%%%%%%%%%%%%%%%%%%%%%%%%%%%%%%%%%%%%%%
    % Sections
    \input{Sections/0_introduction.tex}

    \input{Sections/1_related_work.tex}

    \input{Sections/2_dataset.tex}

    \input{Sections/3_object_pose.tex}
    \input{Sections/4_conclusions.tex}
    %%%%%%%%%%%%%%%%%%%%%%%%%%%%%%%%%%%%%%%%%%%%%%%%%%%%%%%%%%%%%%%%%%%%%%%%%%%%%%%%
    \bibliographystyle{ieeetr}
    \bibliography{bibliography}
\end{document}

%% file: acronyms.tex
\begin{acronym}[ransac]
  \acro{LbD}{learning by demonstration}
  \acro{RL}{reinforcement learning}
  \acro{SVM}{Support Vector Machine}
  \acro{DoF}{degrees-of-freedom}
  \acro{CAD}{computer-aided design}
  \acro{ROI}{regions of interest}
  \acro{MRF}{Markov random fields}
  \acro{ECV}{early cognitive vision}
  \acro{IADL}{instrumental activities of daily living}
  \acro{CDR}{cognitive developmental robotics}
  \acro{DR}{Domain Randomisation}
  \acro{2D}{two-dimensional}
  \acro{3D}{three-dimensional}
  \acro{RANSAC}{random sample consensus}
  \acro{RGB-D}{red-green-blue depth}
  \acro{GPU}{graphics processing unit}
  \acro{RGB}{red-green-blue}
  \acro{ICP}{iterative closest point}
  \acro{IFR}{International Federation of Robotics}
  \acro{CNN}{convolutional neural network}
  \acro{KB}{knowledge base}
  \acro{MSE}{mean square error}
  \acro{MLN}{Markov logic network}
  \acro{XAI}{explainable artificial intelligence}
  \acro{MC-SAT}{model-constructing satisfiability calculus}
  \acro{WCSP}{weighted constraint satisfaction problem}
  \acro{MAP}{Maximum--Likelihood}
  \acro{O-CNN}{octree-based convolutional neural networks}
  \acro{OACs}{object-action complexes}
  \acro{CAD}{computer-aided-design}
  \acro{ROC}{receiver operating characteristics}
  \acro{AUC}{area under the curve}
  \acro{MCMC}{Markov chain Monte Carlo}
  \acro{FOL}{first-order logic}
  \acro{DMP}{dynamic movement primitive}
  \acro{M-RCNN}{mask RCNN}
  \acro{SAGAT}{self-assessment of grasp affordance transfer}
  \acro{OTP}{object transfer point}
  \acro{CHSS}{Chest Heart and Stroke Scotland}
  \acro{ALS}{amyotrophic lateral sclerosis}
  \acro{SRL}{statistical relational learner}
  \acro{WCSP}{weighted constraint satisfaction problem}
  \acro{ANOVA}{analysis of variance}
  \acro{ARAT}{action research arm test}
  \acro{ADL}{activities of daily living}
  \acro{AR}{augmented reality}
  \acro{DL}{Deep Learning}
  \acro{RGB-D}{Red Green Blue-Depth}
  \acro{AUV}{Autonomous Underwater Vehicle}
  \acro{UVMS}{underwater vehicle manipulation system}
  \acro{I-AUV}{intervention autonomous underwater vehicle}
  \acro{AAE}{Augmented Autoencoder}
  \acro{CNN}{Convolutional Neural Network}
  \acro{AE}{Autoencoder}
  \acro{VOC}{Visual Object Classes}
  \acro{CoU}{Complement over Union}
  \acro{ADD}{Average distance to the model point}
  \acro{ADI}{Average distance to the closest model point}
  \acro{IOU}{Intersection over Union}
  \acro{BOP}{Benchmark for 6D Object Pose Estimation}
  \acro{6D}{6 Degrees of Freedom}
  \acro{mAP}{mean Average Precision}
  \acro{SGD}{Stochastic Gradient Descent}
  \acro{PnP}{Perspective'n'Point}
  \acro{FPS}{Frames Per Second}
  \acro{SO(3)}{three-dimensional rotation group}%{Special Orthogonal Group of 3 Dimensions}
  \acro{LiDAR}{light detection and ranging}
  \acro{Sonar}{sound navigation ranging}
  \acro{YOLO}{You Only Look Once}
\end{acronym}

%% file: Sections/0_introduction.tex
\vspace{-0.3cm}
\section{Introduction}\label{sec:intro}

    % Point 1: Introduce why underwater manipulation is needed: hazardous environment and plenty of unknown
    The hazardous nature of underwater environments makes it challenging and critically dangerous for humans to conduct certain underwater intervention tasks.
    To mitigate this, underwater robots have been widely adopted for intervention tasks during the last decade~\cite{palomer20183d}.
    A robotic system used for intervention not only requires a mechanised manipulator arm to interact with nearby objects, but also an {\em object pose estimation} system to track and understand the surroundings.
   
   % Point 2: why is it so difficult to achieve autonomous underwater manipulation
   For an underwater {\em object pose estimation} system to extract the pose of the surrounding objects reliably, it must cope with changes in light conditions, blurriness, and water visibility.
   % Point 3: focus on the sensor input and how expensive 3D sensors are, using a cheap RGB camera
    Employing cutting-edge sensors in complicated setups eases the gathering of cleaner \ac{3D} data. Examples of underwater imaging sensors include sonars and lasers; the former can be cluttered, whereas the latter suffers from distortion on the reconstructed cloud proportional to the sensor motion during the scan~\cite{palomer20183d,sakai1998underwater,shu2004research,czajewski1999development}. These underwater sensing constraints suppose a remarkable challenge for underwater object tracking and intervention~\cite{sanz2015multipurpose}.
   
   %%%%%%%%%%%%%%%%%%%%%%%%%%%%%%%%%%%%%%%%%%%%%%%%%
    \begin{figure}[thb!]
      \centering
      \includegraphics[width= 8.5cm]{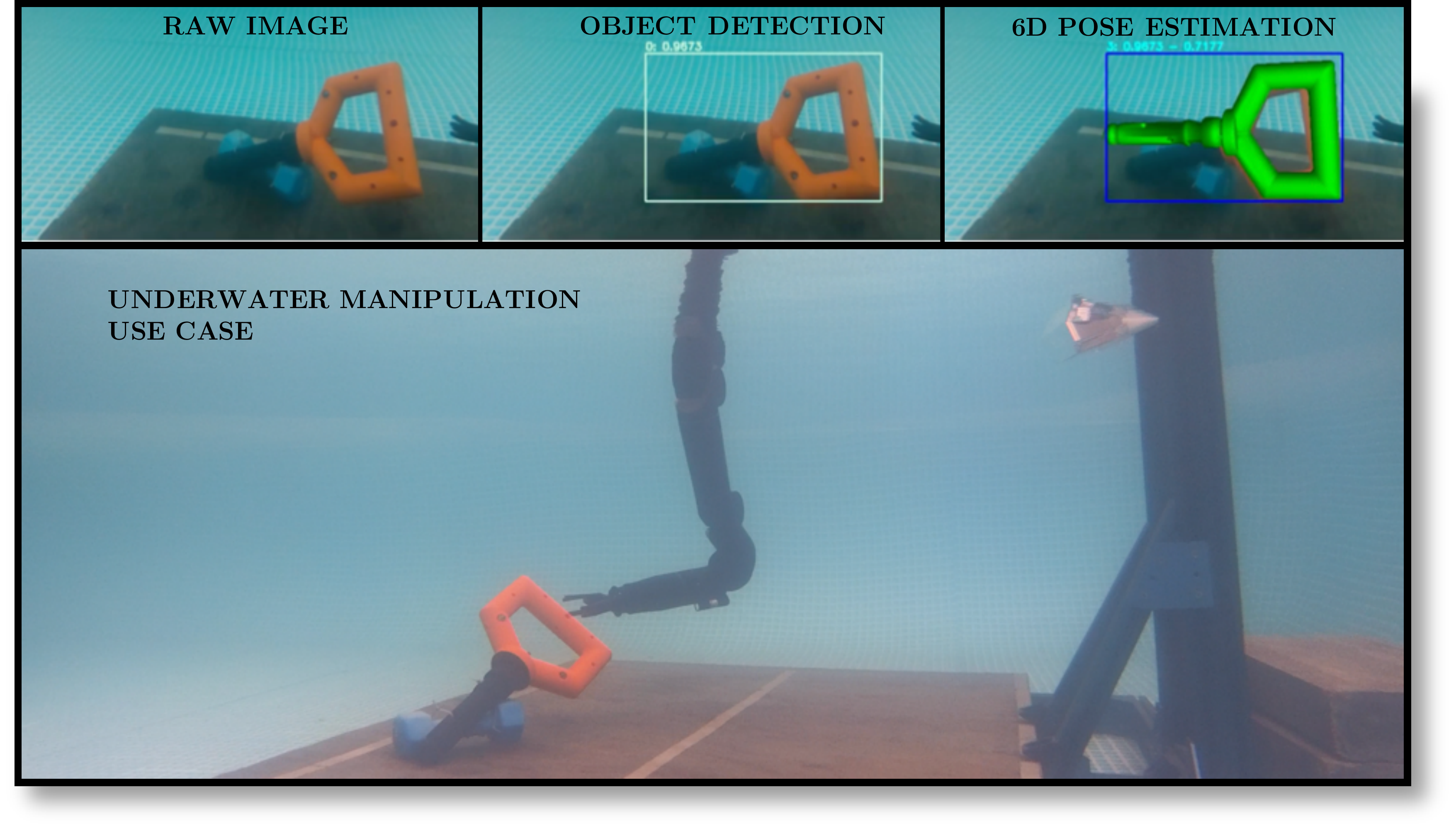}
      \vspace{-0.3cm}
      \caption{Illustration of real-time underwater object pose detection. On top, our processing flow: 1) the RGB as input, 2) detect the object and 3) estimate the object pose. Below, example of our pipeline in an underwater manipulation setting.
      \label{fig:intro} \vspace{-0.7cm}}
    \end{figure}
    %%%%%%%%%%%%%%%%%%%%%%%%%%%%%%%%%%%%%%%%%%%%%%%%%
   
    % Point 4: current methods based on vision and normal sensory
    The lack of on-dry-like \ac{3D} sensing has pushed the advancements of computer vision techniques useful for segmentation and localisation underwater \cite{islam2020semantic,yu2022udepth} as well as \ac{RGB} imagery and \ac{DL} algorithms \cite{joshi2020deepurl,billings2021towards,jeon2019underwater} for known targets.
    %pose estimation. 
    Current methodologies for object pose estimation rely on the prior knowledge of the \ac{CAD} model, which has been proven successful in vehicle tracking~\cite{joshi2020deepurl} and tool detection for intervention applications~\cite{billings2019silhonet,jeon2019underwater}.
    Although the promising results in the literature, these are based on datasets and scenarios that target specific applications, thus lacking generalisation.
    
    One of the main challenges for object pose detection underwater is the lack of datasets with variability of underwater environment representations that can potentiate robust learning algorithms. Data collection in uncontrolled underwater environments is considered a laborious and, in some cases, hazardous task. The unavailability of these datasets hinders the training of object pose detection methods that are robust to the low-light and blurry nature of the input data.
    
     % Point 5: our contributions:
    On the grounds of the limitations mentioned above, the contribution of our manuscript is twofold.
    First, we collect and make available a new rich dataset for underwater object pose detection. This dataset consists of randomised simulated underwater scenarios, as well as $10$ different real-world underwater and dry scenes across two countries (Italy and UAE). The newly available dataset has \ac{RGB} and paired depth images with the corresponding \ac{CAD} models for the objects of interest\footnote{Our dataset is available at \url{https://bit.ly/3LZYvyJ}\label{ft:dataset}}.
    Second, we present a multi-target underwater pose prediction pipeline that employs a combination of deep \acp{CNN}. The architecture employs \ac{YOLO} version 4 \cite{Yolo} for its proven robustness in detecting object instances in underwater \ac{RGB} images~\cite{chen2021underwater}, and an \ac{AAE} \cite{AAE5} to lift the object's pose estimate to 6D while handling symmetries and partial occlusions. Our method trains in simulation and estimates the target pose in various real environments more efficiently than state-of-the-art methodologies that offer an end-to-end solution to model-based pose estimation. We evaluate the reliability of our proposal using the \ac{BOP} toolkit~\cite{hodan2018bop}. We also compare the obtained poses with collected ground truth data and analyse the usability of the method with a robotic arm in a peg-in-a-hole underwater scenario, as shown in Fig.~\ref{fig:intro}.

%% file: Sections/1_related_work.tex
\section{Related Work}\label{sec:related_work}

%%%%%%%%%%%%%%%%%%%%%%%%%%%%%%%%%%%%%%%%%%%%%%%%%%%%%%%%%%%%%%%%%%%%
% A major hustle for underwater object pose estimation is the clarity of visual RGB sensory input. 
A major challenge for underwater object pose estimation is the clarity of sensory input. %Imagery suffers from colour degradation and light scattering.} Alternatively, 3D perception underwater based on sonars, lasers and customised stereo cameras, also faces challenges obtaining clear data~\cite{shu2004research,sakai1998underwater,aldhaheri2022underwater}. 
While imagery suffers from colour degradation and light scattering, 3D perception based on sonars, lasers and customised stereo cameras struggles to acquire accurate and neat point clouds~\cite{shu2004research,sakai1998underwater,aldhaheri2022underwater}.
The lack of on-dry-like 3D sensing motivates underwater object pose estimation to remain an open research topic.

\subsection{Datasets}
    % \cite{palomer20183d} - custom laser \cite{penalver2015visually,perez2015robotic} - visually guided manipulation (either LEDs or aruco codes)
    Underwater datasets for object pose detection are labour-intensive to collect.
    %given the complexity of the environment.
    This section overviews available datasets useful in dry and underwater environments for object tracking and manipulation from multiple types of sensors.
    
    Assuming ideal conditions and perfect acquisition of \ac{RGB} images from a stereo pair, the corresponding depth information can be computed.
    However, in reality, ideal conditions are rarely met. For this reason, \cite{palomer20183d} develops its own underwater \ac{3D} laser scanner. The resulting point cloud is used to guide an eight \ac{DoF} fixed-base manipulation system to follow pre-recorded trajectories in real time. In a subsequent demonstration, an \ac{UVMS} uses the same sensor to autonomously pick up an object from the bottom of a water tank.
    Other methodologies resort to using commercial sensors to overcome the object detection challenge while compensating for the object pose by having a prior of the object shape. For example, \cite{billings2020silhonet} presented an underwater dataset called UWHandle of three types of graspable handles collected from a natural seafloor environment. In \cite{billings2020silhonet}, they project the known handle models on the \ac{ROI} detected from a fish-eye camera and test the pose accuracy with AprilTags fiduciary \cite{olson2011apriltag} markers. \cite{jeon2019underwater} also proposes a \ac{3D} model-based method with a subset of objects in a controlled water tank setting. While these methodologies propose underwater pose detection, their accuracy is limited to specific environments which limits their extension to manipulation or object tracking in real scenarios.

\subsection{6D Pose Estimation}
% \cite{billings2020silhonet} - SilhoNet
 
    Spatial awareness from obtaining \acs{6D} pose estimation between a robot and a reference is essential for successful tracking or manipulation tasks. Limited literature exists for object pose estimation in underwater environments.
    
    % \cite{billings2020silhonet} besides collecting an underwater dataset for object pose detection of known handles, it proposes a method that detects poses of objects under controlled partial occlusions.
    Authors in \cite{billings2020silhonet} proposed a method to detect poses of objects under controlled partial occlusions. Their approach is to regress the \ac{3D} poses from monocular silhouettes predicted by a \ac{CNN} pipeline that uses an associated occlusion mask of the known object \ac{3D} model with a transformation vector. This architecture required data acquisition in real environments to estimate the object pose while knowing the target object. Also knowing the \ac{3D} model in advance, authors in \cite{joshi2020deepurl} propose a real-time \acs{6D} relative pose estimation of an \ac{AUV} from a single image. This approach uses underwater simulated scenario photos for training. In order to create synthetic images for training, an image-to-image translation network is used to close the gap between rendered and real images. The suggested method predicts the pose of an AUV from a single \ac{RGB} image that corresponds to the 8 corners of the AUV's \ac{3D} model. The resilience and accuracy of the suggested technique are demonstrated in real-world underwater environments with different cameras. While these works motivate the development of underwater object pose estimation methods, they do not cope with uncontrolled object occlusions and blur caused by marine life and diverse conditions in the underwater environment.
    
    %Methodologies such as \cite{billings2020silhonet} require data acquisition in real environments to estimate the object pose while knowing the target object. While these previous works motivate the development of underwater object pose estimation methods, they do not consider challenges such as \textcolor{blue}{object occlusions and blur caused by marine life and diverse conditions in the underwater environment.}
    % dynamic changes in the environment, such as currents, turbidity and moving objects with marine life and debris. 

    Contrary to existing underwater pose estimation methodologies, we make available online a dataset consisting of four different objects across simulated and real environments considering scenes with the objects partly occluded and diversity of backgrounds ranging from homogeneous colours to vegetation.
    % dynamically changing environments with turbidity and marine life.
    To estimate the object pose, instead of learning an explicit mapping from input images to object poses we propose a pipeline that provides an implicit representation of object orientations defined by samples in a latent space. Thus allowing to train in simulated and generalise to real underwater scenes. Moreover, as explored in Section~\ref{sec:methodology}, our proposal shows to be robust to lack of textural surface and symmetrical object geometries.

%% file: Sections/2_dataset.tex
\section{Our Dataset}\label{sec:dataset}

    % Dataset description
    This work pursues an underwater multi-object pose estimation for which the training data needs to be diverse and accurate. 
    % However, when learning object pose estimation for underwater robotic intervention, one of the greatest challenges is the lack of datasets. Contrary to the variety of existent \ac{RGB} and depth object datasets for dry environments \cite{hinterstoisser2011gradient,xiang2017posecnn,lin2014microsoft} there are no object datasets for underwater intervention settings.
    However, contrary to the vast \ac{RGB} and depth object datasets for dry environments \cite{hinterstoisser2011gradient,xiang2017posecnn,lin2014microsoft}, there is a lack of datasets for underwater intervention settings.
     We present a new dataset to push the state-of-the-art in underwater object recognition and pose estimation. Such dataset is then employed in Section~\ref{sec:methodology} to compute object poses with our framework.% presented in Section~\ref{sec:methodology}.

    Our dataset contains \ac{RGB} annotated frames of four different object categories seen from different points of view alongside the objects' \ac{3D} models. The chosen object categories represent common use cases for underwater intervention tasks, such as maintenance or object recovery. The dataset is intended to continue growing with various object representations useful for underwater intervention purposes and to encourage the creation of robust underwater object pose estimation methods.

    % Dataset collection
    \subsection{Data Collection and Generation \label{sec:data_collection}}
        \subsubsection{Data generation}
            in order to test our 6D pose estimation framework proposed in Section~\ref{sec:methodology}, we designed four different objects commonly used for underwater intervention tasks. These objects represent symmetrical, asymmetrical and texture-less shapes. Fig.~\ref{fig:dataset} illustrates our object categories. Using Unity and the \ac{CAD} model of our objects, we generate a total of $33{,}920$ simulated scenes of dimension $640\times480$ with different homogeneous and heterogeneous backgrounds, including object occlusions and various illuminations. Fig.~\ref{fig:dataset} showcases some of our simulated scenes. 
        \subsubsection{Collection in real environments\label{sss:real_environments}}
\begin{figure}[tb!]
                % \centering
                 \centering \includegraphics[width=8.6cm]{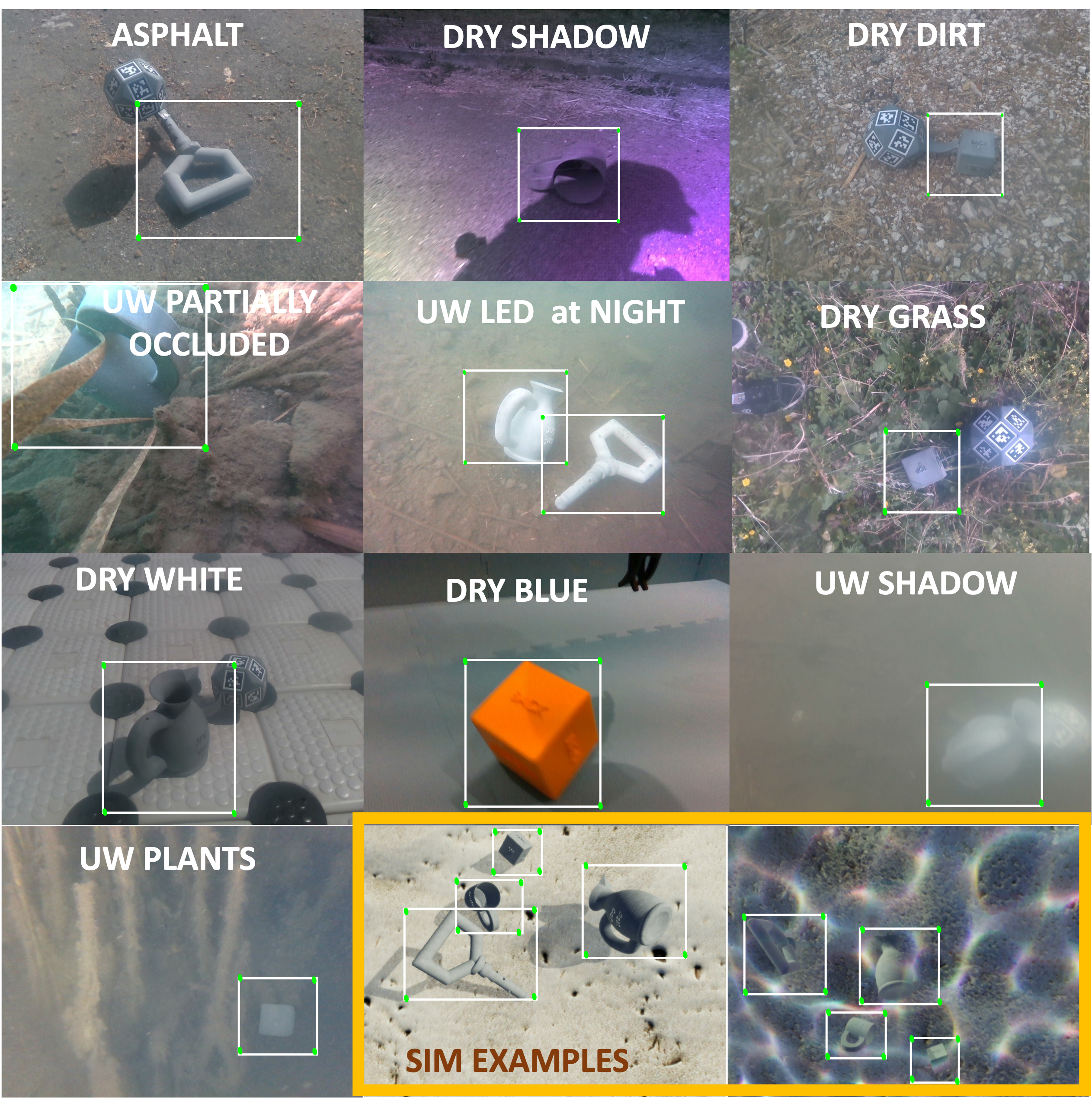}
                \caption{Example of our dataset, including real (white font) and simulated scenes (black font). \textit{UW describes underwater.}
                \label{fig:dataset}}
            \end{figure}

            we 3D printed our objects to harvest data in real scenes. The dataset is collected using a d455 realsense inside a waterproof container. The camera simultaneously records both, \ac{RGB} and depth images at ${640\times480}$ resolution at 30 frames per second. For the development of the framework in Section~\ref{sec:methodology}, we solely use the \ac{RGB} data for training. While we do not use the depth information for our method, we provide the depth frames in our available dataset for the use of the community.
            Using the previously described camera setup, we record video sequences of each of the four object categories in $10$ different setups, both, individually and using various categorical combinations. The average scene sequence contains $180$ \ac{RGB} frames with corresponding depth data. The different scenes are recorded across two different countries (Italy and UAE), in different real underwater and dry environments. For our underwater scenes, we consider environments, such as partial object occlusions from marine life, shadows and changing lighting conditions, low visibility at night coped with LED illumination and heterogeneous backgrounds. Fig.~\ref{fig:dataset} shows examples of our real underwater scenarios (captioned `UW' in white font). Per recording, we freely rotate the object inside the camera field of view at varying distances as far as $3m$ from the target, as bounded by hardware limitations. 
            %------------------------------

    % Dataset annotation and processing
    \subsection{Object Recognition Annotation and Processing}
        \subsubsection{Annotation}
            given the diversity of environments in our dataset we need an efficient yet high-quality annotation pipeline to define our objects of interest in the images. We annotated a total of $87{,}100$ real \ac{RGB} images for \ac{2D}. Examples of the bounding boxes are shown in Fig.~\ref{fig:dataset}. Our annotation process consists of three stages. First, a human annotator labels approximately $1\%$ of the data across scenes, including partial views of the objects. Second, we designed an auto-labelling tool based on \ac{YOLO}v4 \cite{Yolo} to create the bounding boxes for the rest of the data based on the $1\%$ of manually annotated data. Finally, three different human annotators checked the created bounding boxes in a sequential fashion to ensure the quality of the annotated region of interest.
        \subsubsection{Dataset Statistics}
\begin{figure}[tb!]
                % \centering
                 \centering \includegraphics[height=5.5cm,width=8cm]{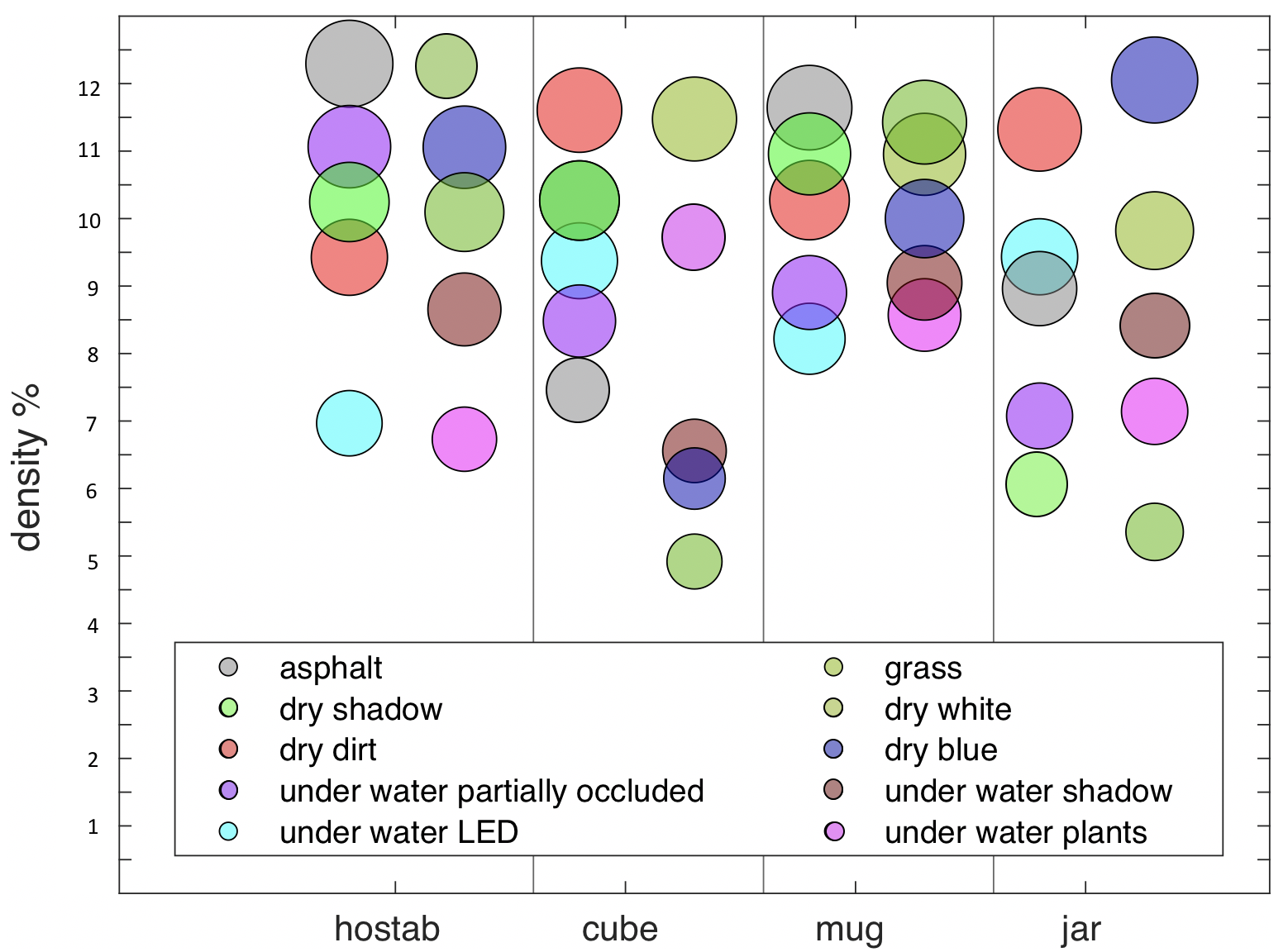}
                \caption{Dataset collection density across scenes.
                \label{fig:density}}
            \end{figure}

\begin{figure}[hb!]
                \centering
                \subfloat[Hotstab\label{fig:hotstab}]{
                    \centering
                \includegraphics[width=4.2cm]{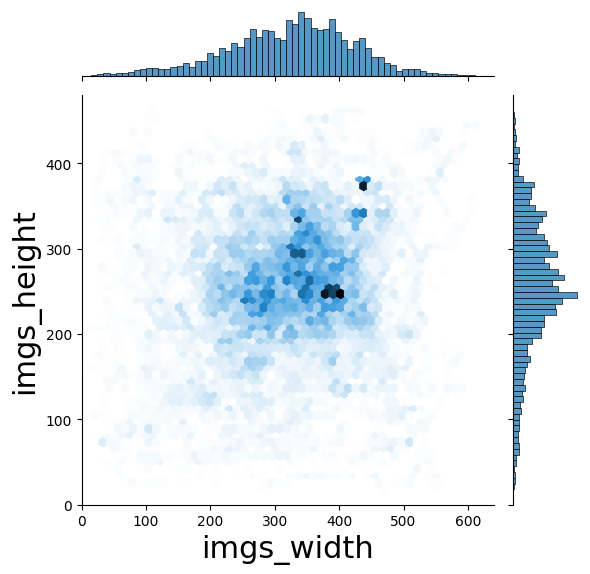}
                }
                \subfloat[Cube\label{fig:cube}]{
                    \centering    \includegraphics[width=4.2cm]{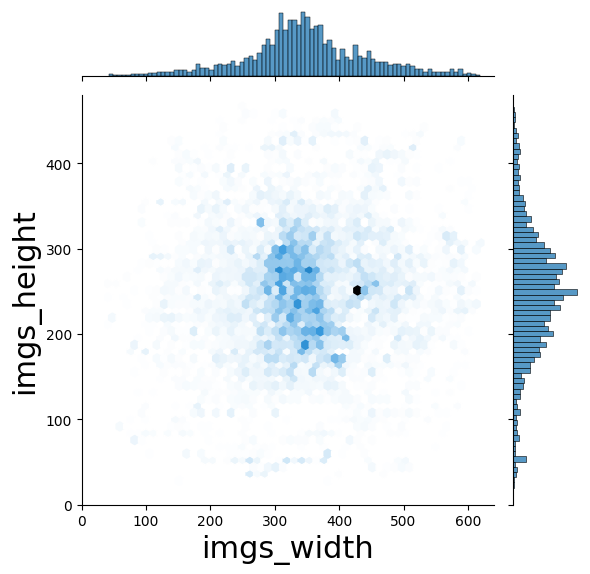}
                } 
                \vspace{-0.4cm}
                \subfloat[Mug\label{fig:cup} ]{
                    \centering
                        \includegraphics[width=4.2cm]{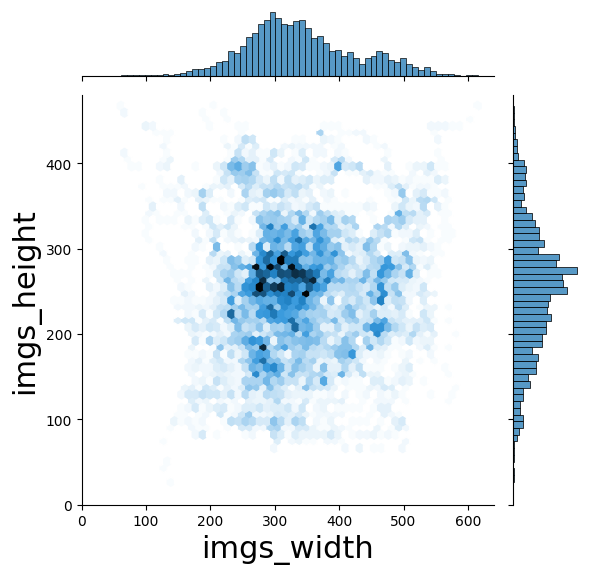}
                }
                \subfloat[Jar\label{fig:jar}]{
                    \centering
                \includegraphics[width=4.2cm]{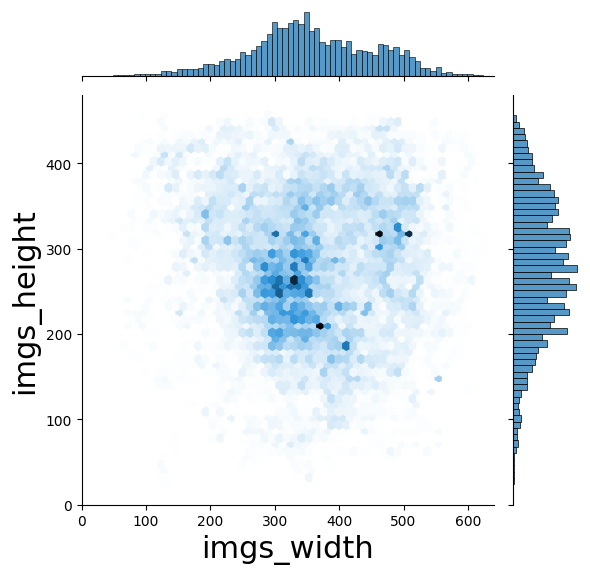}
                }
                    
                \caption{Annotations representing object locations in the image.} \label{fig:histogramLabels}
            \end{figure}

        %check: bounding boxes size, class balance
            we consider an important feature of our dataset the variability of scenes, particularly the contribution of underwater imagery. Fig.~\ref{fig:density} shows the four objects across our $10$ scenes and the percentage of \ac{RGB} images per scene. As seen from the data density, the collected data is spread across different settings. The real underwater collected images represent $42.5\%$ of our dataset, being collected in $4$ different underwater conditions. The other $57.5\%$ are image sequences collected from $6$ on dry scenes.
            %------------------------------
           %------------------------------
            Fig.~\ref{fig:histogramLabels} shows the location distribution of the centre of our annotated bounding boxes on the \ac{RGB} images. As shown in the plots, there are some concentrations at the centre of the ${640\times480}$ image, with some bounding boxes located across most of the pixels.
            
           %------------------------------
            %----------------------

    \subsection{Ground Truth Extraction for Pose Estimation \label{ss:bundle_gt}}
\begin{figure}[tb!]
                % \centering
               \centering \includegraphics[width=6cm]{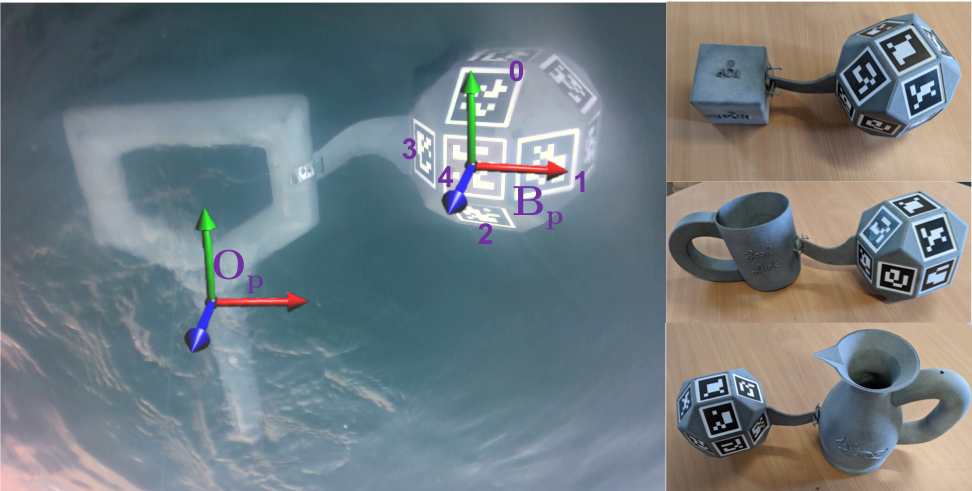}
                \caption{On the right, example of the annotated ground pose estimation on the hotstab. On the left, rest of the objects.
                \label{fig:ground_truth}}
            \end{figure}

        %ground truth extraction
        Exclusively for sim-to-real evaluation purposes, to obtain the ground truth of the object pose, we attached an AprilTag~\cite{olson2011apriltag} bundle to the target objects. Thus, the bundle has a fixed map to the centre of geometry of the object. As illustrated in Fig.~\ref{fig:ground_truth}, given that we know the object base frame by its \ac{3D} design, $O_p$, and we can detect the bundle pose, $B_p$ we calculate offline the transformation between the object and the bundle, $T^O_B$. Using this information we are able to extract the ground pose and estimate the error of our detection (detailed in Section~\ref{ss:error_estimation}). %An example is illustrated in Fig.~\ref{fig:ground_truth}.

         %------------------------------
           %------------------------------
    
     In general our collected dataset displays features that promote detection robustness across different scenes, including dry and underwater environments, as well as a high-quality set of annotations for future proposals on learning methodologies.

        %poses with apriltag bundle

%% file: Sections/3_object_pose.tex
\section{Object Pose Estimation\label{sec:methodology}}
\begin{figure}[tb!]
        \centering
        \includegraphics[width=8.6cm]{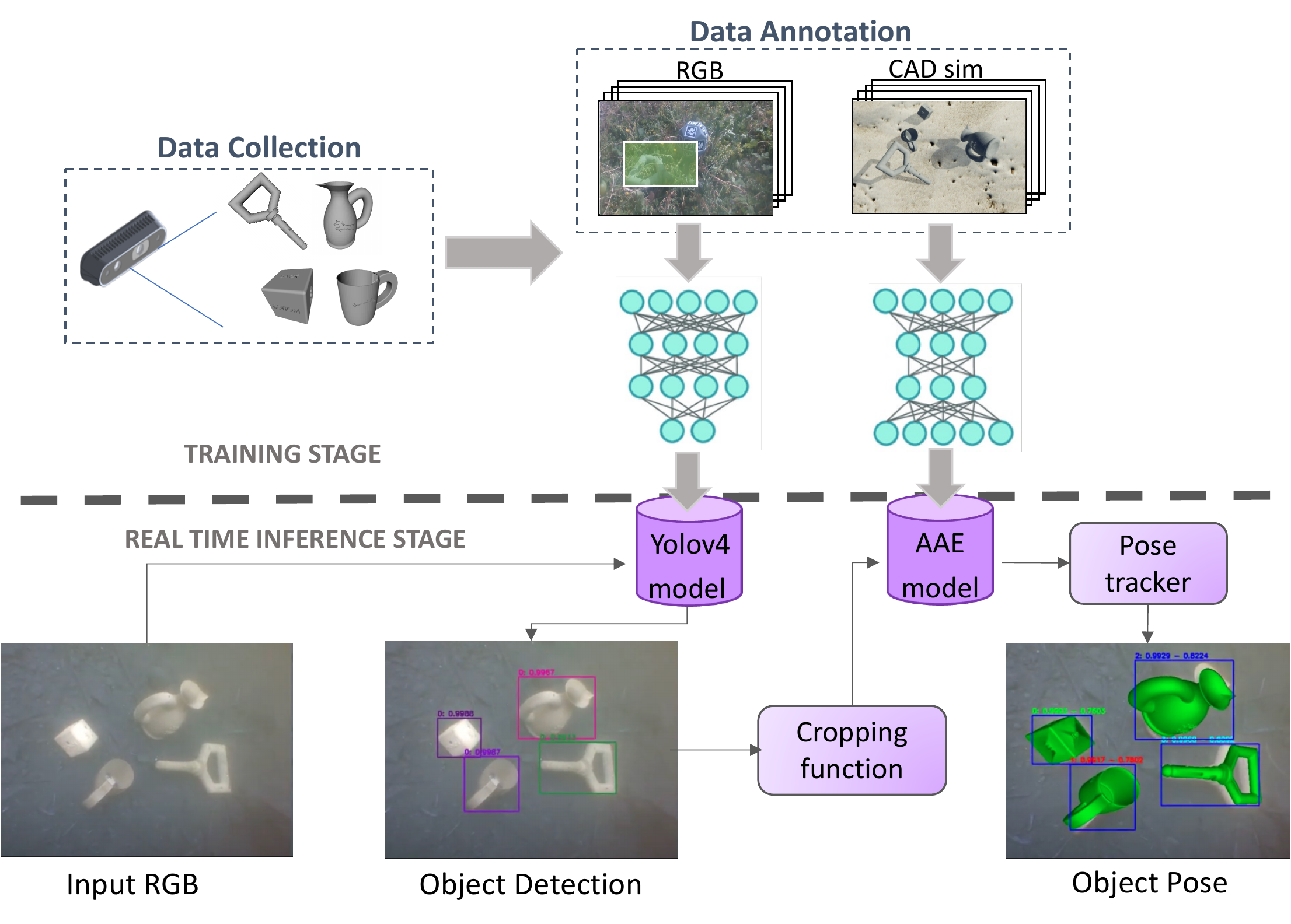}
        \caption{Given an \ac{RGB} image as input, the first step consists of \ac{2D} detection and classification with multi-object \ac{YOLO}v4. Given cropped images and their classes, they become the input of one of the four \ac{AAE} training, one for each object. As output, each training returns a rotation matrix $R$ and a translation matrix $t$.}
        \label{fig:diagram}
    \end{figure}

    %As previously stated, obtaining reliable \ac{3D} data underwater is difficult given the limitations on reliable, and when in existence, expensive hardware.
    Obtaining reliable \ac{3D} data underwater is difficult given the environment and sensing limitations.
    Given the aforementioned limitations, we propose a pipeline composed of \ac{2D} object detection and \acs{6D} pose estimation on the \ac{RGB} image. A diagram of our pipeline is illustrated on Fig.~\ref{fig:diagram}. On the detected object bounding box, we then estimate the object \acs{6D} pose. Namely, we create a pipeline based on \ac{YOLO}v4 for \ac{2D} object detection (as detailed in Section~\ref{ss:yolov4}) and the \ac{AAE} for \acs{6D} pose estimation (as detailed in Section~\ref{ss:AAE}). The pipeline is multi-object because the \ac{YOLO}v4 is multi-object and selects the correspondent object's \ac{AAE} training (one for each class).
    Our pipeline is designed to find the best trade-off between computational efficiency and accuracy on real-time input while proving to be robust for detection and pose estimation in dry and underwater environments.
    Opting for this pipeline offers a significant advantage because it can be completely trained in simulation and only requires labelled data for the \ac{YOLO}v4, while generalising on real scenes (see Section~\ref{ss:error_estimation}). Our dataset is readily available and includes \ac{RGB} images and \ac{2D} labels containing bounding box information of the depicted objects. Conversely, no dataset is necessary for AAE since it self-builds from the \ac{CAD} model. Nevertheless, synthetic labelled \acs{6D} datasets and real labelled \acs{6D} datasets were produced and used during the validation of our method (see Sections~\ref{sssec:baseline},~\ref{sssec:benchmark},~\ref{ss:error_estimation}). The significance of \acs{6D} labelled datasets extends to being applicable to almost all other underwater pose estimation techniques that necessitate \acs{6D} labels. Moreover, differently from similar methodologies for \acs{6D} pose estimation, our proposed pipeline proves to be unaffected by the object perception point-of-view, self occlusions, and symmetric geometries as illustrated in Section~\ref{ss:evaluation}.
    
    \subsection{2D Object Detection \label{ss:yolov4}}
        For \ac{RGB} based pose detection methods, accurate \ac{2D} detection is essential for the subsequent pose estimation.
        To train object detection we consider various underwater scenes to enrich the collection of features, as detailed in Section~\ref{sec:data_collection}.
        % Given the complexity of the underwater scene, we consider different cases, such as blurring effects and lighting changes as the object goes to greater depths underwater.
        
        In the context of deep learning, \ac{YOLO}v4 has demonstrated to efficiently perform in a heterogeneous environment, with an excellent trade-off between accuracy and latency \cite{DBLP:journals/corr/abs-2004-10934}. 
        %To bridge the domain gap between sim to real, we use both, real and synthetic data (as introduced in Section~\ref{sec:dataset}) in the training stage.
        The training used only synthetic data. As detailed in Section~\ref{sec:data_collection}, our synthetic data considers different backgrounds, light colours and directions, in order to overcome issues of the underwater scenario. Moreover, \ac{YOLO}v4 applies data augmentation to the original data. \ac{YOLO}v4 uses a saturation of $1.5$, exposure of $1.5$, hue $0.1$, and a mosaic effect: it combines multiple images into a single image.
        We use \ac{YOLO}v4 architecture as originally proposed in \cite{DBLP:journals/corr/abs-2004-10934} to extract a bounding box with the object features in real-time.
       
        Up to this stage in our pipeline, we have a detected object and the representing class as the output from \ac{YOLO}v4. At this point, we still need to estimate the detected object 6D pose.
        
    \subsection{6D Pose Estimation \label{ss:AAE}}
            
        After \ac{YOLO}v4, a method for 6D prediction should be chosen. A major issue that usually affects \acs{6D} pose estimation is pose ambiguity. Usually, these ambiguities arise due to symmetries or self-occlusion of the objects. For example, when the cup's handle is occluded, the cup's pose is ambiguous.
        Therefore choosing a pose ambiguity invariant approach has been revealed to be essential. In addition, given the underwater scenario, the method should be robust to light changes and blurring effects in real-time. For these reasons, an \ac{AE} structured model is chosen: \ac{AAE} \cite{AAE5}. On the contrary, alternative approaches \cite{33,39,tremblay2018deep}, involve regressing local 2D-3D correspondences between the image and CAD model, followed by the application of a Perspective-n-Point (PnP) algorithm to derive the object's 6D pose. However, these methods present two limitations in our scenario. Firstly, they require clear images to distinguish visual features, which poses a challenge in our underwater environment. Secondly, these methods are not pose invariant, and as such, are not suitable for our scenarios where 3 out of 4 objects are symmetrical. 
        On the other hand, \ac{AAE} was proposed by \cite{AAE5} as a symmetry invariant approach. This is due to the property of representing the orientation not on a fixed parametrisation, but on the appearance of an object. Our specific underwater dataset, achieves higher performance scores than other widely used models, as shown in Section~\ref{ss:metrics_evaluation}, demonstrating to be robust to underwater scenarios. Furthermore, \ac{AAE} is suitable to real-world use case requirements, as shown in Table~\ref{tab:times} where inference times are computed.
     
        We explored different methods, as detailed in Section~\ref{ss:evaluation}, and found out that \ac{AAE} works better in our environment because of the properties outlined above.
        Since rotations live in a continuous space, it seems natural to directly regress a fixed rotation in the \ac{SO(3)}, like in EfficientPose \cite{efficientpose}. However, pose ambiguities and representational constraints can introduce convergence issues \cite{39}. \ac{AAE} solves this problem by discretising the \ac{SO(3)} space and then moving the problem from regression to classification. 
        Furthermore, an advantage of \ac{AAE} is that it does not depend on real data collection and annotations for training. Instead, in the training stage, \ac{AAE} creates its own dataset by randomising object poses, taken from the \ac{CAD} model, and \ac{VOC} background real images, to bridge the simulation to real images gap. 
        To guarantee the \ac{AAE} performance on dry and underwater scenarios, we (i)~add background images from our collected dataset in real scenarios (detailed in Section~\ref{sss:real_environments}), and (ii) customise the data augmentation for training stage.
        \ac{AAE} applies data augmentation at the pre-processing stage when it generates its dataset, allowing the method to be trained solely in simulated data. Table~\ref{tab:hyperparam} reports the \ac{AAE} data augmentation hyperparameters.  
                
        A pose of a \ac{3D} object in this proposal is represented by the $4\times4$ matrix $\mathbf{P}=[\mathbf{R,t; 0},1]$, where $\mathbf{R}$ is the rotation represented by a $3\times3$ matrix and $\mathbf{t}$ is the translation represented by a $3\times1$ vector. $\mathbf{P}$ transforms a 3D point $x_m$ in the model coordinate system to a 3D point $x_c$ in the camera coordinate system: $\mathbf{P}[x_m;\mathbf{1}] = \mathbf{R }x_m+\mathbf{t }= [x_c;\mathbf{1}]$ 
               
        \ac{AAE} is based on \ac{AE} structure, aiming to obtain an image representation in a low-dimensional Euclidean space.
        In detail, first, \ac{AAE} applies a random augmentation $f_{aug}$ to input $x$ and reconstructs the original image. Then, an encoder-decoder training reconstructs the original input. The parameters are learned during the backpropagation phase, based on the per-sample loss: $l = \sum_{i\in D} \Vert x_i - \hat{x}_i \Vert_2$. After this initial training, a \textit{codebook} is created by generating a latent representation $z_i \in \mathbb{R}^l$ of each one of the $n$ object views, and their correspondent $\mathbf{P}_i$ matrices. The \ac{AAE} training must be done for each one of the objects in the dataset. During the test phase, \ac{AAE} is preceded by the \ac{2D} detector and receives as input the already detected and cropped image. The image goes through the encoder which gives its latent space features.
        Then, the cosine similarity is computed between the input latent representation code and all codes from the codebook. The highest similarity is chosen and the corresponding rotation matrix from the codebook is returned as \ac{3D} object orientation.

        For the entire pipeline, five training are needed: one multi-object \ac{YOLO}v4 and four different \ac{AAE}, one for each object. Despite this, the pipeline is considered multi-object since in inference, given an image, is able to reconstruct the pose of each observable object in time explained in Table \ref{tab:times}. One limitation of \ac{AAE} is that the number of objects in a single scene affects the inference time. However, in the case of four objects the achieved performance respects predefined time constraints.
        Deep Neural Networks performance, as \ac{AAE}, is dependent on hyperparameters that determine the network structure. Finding the right hyperparameters is fundamental to ensuring good performance. 
        \ac{AAE} relies on many hyperparameters \cite{AAE5}, which are divided into two groups: those modifying the structure and optimization of the network (such as learning rate, latent space dimension, batch normalization), and those acting on the data augmentation (such as occlusion percentage, inversion, multiplication, drop, Gaussian blurring addition). 
        % number of hyperparameter configurations, it is not possible to explore all of them, so heuristics are used to choose only the most promising configurations.
        Given the large hyperparameter space, we use heuristics to only explore the most promising configurations. In our case, we ran experiments, on a total of $40$ different configurations to find the best hyperparameter setting.
            
            \begin{table}[tb!]
                \centering
                \begin{tabular}{|c|c|c|c|c|}
                    \hline
                    &  Box & Cup & Jug & Hotstab \\
                    \hline
                    \multicolumn{5}{|c|}{Data Augmentation Hyperparameters}\\
                    \hline
                    \scriptsize{Perspective Transform} &  &  &  & \\
                   
                    \scriptsize{Crop And Pad} & \checkmark & \checkmark & & \checkmark\\
                      
                      \scriptsize{Affine}  & \checkmark& \checkmark & \checkmark & \checkmark \\
                        
                      \scriptsize{Coarse Dropout}  & \checkmark & \checkmark & & \checkmark \\
                       
                       \scriptsize{Gaussian Blur} & & & \checkmark &\checkmark \\
                        
                       \scriptsize{Invert} & \checkmark & \checkmark & \checkmark &  \checkmark\\
                    
                    \scriptsize{Multiply} & \checkmark& \checkmark & \checkmark& \checkmark \\
                  
                     \scriptsize{Contrast Normalization} & & &\checkmark & \checkmark \\
                   
                        \scriptsize{Square Occlusion} & $0.6$ & $0.4$ & $0.4$ & $0.4$ \\
                    \hline
                        \multicolumn{5}{|c|}{Architecture-Hyperparameters}\\
                 \hline
                    \scriptsize{Learning Rate} & $2e-4$ & $2e-4$ & $2e-4$ & $2e-4$\\
                        
                        \scriptsize{Optimizer} & Adam & Adam & Adam & Adam \\
                        
                        \scriptsize{Latent Space Dimension} &  $256$ & $256$ & $128$ &  $256$ \\
                       
                        \scriptsize{Epochs} & $50,000$ & $70,000$ & $70,000$ & $70,000$ \\
                        
                        \scriptsize{Batch Size} & $32$ & $32$ & $64$ & $64$\\
                        
                    \hline
                \end{tabular}
                \caption{Hyperparameters chosen for each object's training.}
                \label{tab:hyperparam}
            \end{table}
            
           After different training for architecture network and data augmentation hyperparameters optimization, we choose the best set per each object training (see summary in Table~\ref{tab:hyperparam}).

            % Before choosing this approach, we compared \ac{AAE} also with two state-of-the-art methods, results are shown in Section~\ref{ss:evaluation}.
        
    \subsection{Method Evaluation\label{ss:metrics_evaluation}}

        Our evaluation is threefold. First, using the simulated data per object in our dataset, we provide a baseline of our method's performance using part of the \ac{BOP} metrics \cite{hodan2018bop} and \ac{CoU} (see Section~\ref{sssec:baseline}). 
        Second, using the \ac{BOP} metrics, we benchmark against current literature. We choose \ac{YOLO}-6D \cite{39} and EfficientPose \cite{efficientpose}, as they also are deep learning methods that generalise mappings from images to objects. Nonetheless, as they do not claim to bridge training in simulation to real scenarios performance, the evaluation is done in underwater simulated data (see Section~\ref{sssec:benchmark}). Finally, to test the robustness of our training in simulation and its performance in real scenes, we calculate the error pose using the ground truth (see Section~\ref{ss:error_estimation}). This error estimation provides a baseline for future work to compare our method's performance and dataset.

        For the evaluation in Section~\ref{sssec:baseline} and \ref{sssec:benchmark}, we use three of the widely popular (\!\!\cite{posecnn,efficientpose,39}) \ac{BOP} toolkit metrics:
            \begin{itemize}
                \item \textbf{\Ac{mAP}} is a 2D bounding box detection metric $mAP(C, \Gamma, R) = avg_{c \in C} AP(c, \Gamma, R)$, where $AP(c, \Gamma, R)$ is the area under the curve, $\Gamma$ is the set of thresholds used in \ac{IOU} scores, $c$ is the class, and $R$ is the set of the discretised recall values;
                \item  \textbf{\ac{ADD}} given the model $\mathcal{M}$, the estimated pose $\hat{\mathbf{P}}$ and the ground-truth $\mathbf{\overline{P}}$ $e_{ADD} = avg_{x\in \mathcal{M}} \Vert \mathbf{\hat{P}x}-\mathbf{\overline{P}x} \Vert$;
                \item  \textbf{\ac{ADI}} if the model $\mathcal{M}$ has indistinguishable views, then $e_{ADI} = avg_{\mathbf{x}_{1}\in \mathcal{M}} \min_{\mathbf{x_2} \in \mathcal{M}} \Vert \mathbf{\hat{P}x_1}-\mathbf{\overline{P}x_2} \Vert$.  $e_{ADI}$ yields relatively small errors since it does not consider model point distances, but the distance to the closest model diameter.
            \end{itemize}
            
        Moreover, we compute the \textbf{\ac{CoU}} error, which compares the predicted pose and ground truth masks from the \ac{CAD} model.

        \subsubsection{Dataset Baseline with \ac{BOP}\label{sssec:baseline} given that the benchmark in Section~\ref{sssec:benchmark} is produced with simulated data, we provide a performance baseline per object in our dataset with synthetic images to have a fair and uniform comparison baseline.} For this section, we use $500$ synthetic underwater scenes in various settings as detailed in Section~\ref{sec:data_collection}.
        In our dataset, the box, cup, and hotstab are symmetrical, while the jug is not. To evaluate \ac{AAE}'s performance, we choose $e_{ADI}$ for the symmetrical set and $e_{ADD}$ for the asymmetrical object. Our results with different thresholds of correctness for $e_{ADI}$ are illustrated in Table~\ref{Tab:adi}. Table~\ref{Tab:adi} introduces a baseline for our new underwater dataset. For the jug, we computed the $e_{ADD}$, obtaining $29.2\%$ for  $k_m=0.1$; $56.6\%$ for $k_m=0.2$, and; $68.6\%$ for $k_m=0.3$. Nonetheless, these metrics are not pose-ambiguity invariant. Consequently, the objects may present ambiguous poses. This is due to their symmetrical views. For example, the box is texture-less and has two symmetrical axes, thus being difficult to distinguish between its faces. 
        \begin{table}[!b]
            \centering
            \begin{tabular}{|c||c|c|c|}
            \hline
                Objects  & $k_m=0.1$ & $k_m=0.2$ & $k_m=0.3$ \\
                \hline
                Box  & $21.6\%$ & $51.0\%$ & $60.4\%$\\
                Cup  & $55.0\%$ & $77.8\%$ & $85.8\%$\\
                Jug  & $72.2\%$ & $86.6\%$ & $93.0\%$\\
                Hotstab  & $44.0\%$ & $64.4\%$ & $73.8\%$\\
                \hline
                \textbf{Average Perc.:} & $48.2\%$ & $69.95\%$ & $78.25\%$ \\
                \hline
            \end{tabular}
            \captionof{table}{Recall percentages based on $e_{ADI}$. \label{Tab:adi}} 
        \end{table}
        For these cases, we also calculate the \ac{CoU} on the same set of images. The results are shown in Table~\ref{Tab:cou}. Contrary to the results in Table~\ref{Tab:adi}, \ac{CoU} does not penalise symmetric poses, since it uses only segmentation masks. A clear example is the different pose estimation performance on the box object between $e_{ADI}$ from Table~\ref{Tab:adi} and $e_{COU}$ in Table~\ref{Tab:cou}.

        \begin{table}[b!]
            \centering
            \begin{tabular}{|c||c|c|c|}
            \hline
             \ac{CoU} error  &  $\theta=0.3$& $\theta=0.5$ & $\theta=0.7$\\
            \hline
            Box  & $94.6\%$ & $100\%$ & $100\%$\\
            Cup   & $94.2\%$ & $99.4\%$ & $100\%$\\
            Jug   & $79.6\%$ & $97.8\%$ & $99.6\%$\\
            Hotstab   & $20.2\%$ & $65.6\%$ & $92.6\%$\\
            \hline
            \textbf{Average Percentages:} & $72.15\%$ & $90.57\%$ & $98.05\%$\\
            \hline
        \end{tabular}
        \captionof{table}{Recall percentages based on $e_{COU}$. \label{Tab:cou}}
        \end{table} 

       Additionally, in terms of real-time performance, Table~\ref{tab:times} shows the inference times of our proposal per detection stage and end-to-end pipeline. These results use the same validation set, thus the statistics of $500$ inference times. The first time is discarded to not consider the weight load and initialisation memory levels. Inference times are computed on a Xavier AGX, Jetpack $5.0.2$.
           
            \begin{table}[tb!]
                \centering
                \begin{tabular}{|m{0.8cm}m{0.3cm}|m{0.5cm}|m{0.6cm}|m{0.6cm}m{0.6cm}|}
                %{|cl|rr|rr|rr|}
                \hline
                    \multicolumn{1}{|l}{}                                &     & \multicolumn{1}{c|}{\tiny{\ac{YOLO}v4}}                            & \multicolumn{1}{c|}{\tiny{AAE}}                               & \multicolumn{2}{c|}{\tiny{END TO END}}                                 \\ \cline{3-6} 
                    \multicolumn{1}{|l}{}                                &     & {\scriptsize{FPS}}     & {\scriptsize{FPS}}   & \multicolumn{1}{l|}{\scriptsize{latency (ms)}}      & \multicolumn{1}{l|}{\scriptsize{FPS}} \\ \hline
                    \multicolumn{1}{|c|}{\multirow{3}{*}{\tiny{single object}}} & \tiny{min} & {\scriptsize{3.57}}                     & {\scriptsize{8.14}}   & \multicolumn{1}{r|}{\scriptsize{403.13}}   & \scriptsize{2.48}                     \\ \cline{2-6} 
                    \multicolumn{1}{|c|}{}                               & \tiny{max} & {\scriptsize{2.93}}                   & {\scriptsize{6.37}} & \multicolumn{1}{r|}{\scriptsize{497.78}}   & \scriptsize{2.01}                     \\ \cline{2-6} 
                    \multicolumn{1}{|c|}{}                               & \tiny{avg} & {\scriptsize{3.42}}   & {\scriptsize{7.44}}                     & \multicolumn{1}{r|}{\scriptsize{426.51}}   & \scriptsize{2.34}                     \\ \hline
                    \multicolumn{1}{|c|}{\multirow{3}{*}{\tiny{multi objects}}} & \tiny{min} & {\scriptsize{3.52}}                     & {\scriptsize{1.83}}   & \multicolumn{1}{r|}{\scriptsize{830.61}}   & \scriptsize{1.20}                     \\ \cline{2-6} 
                    \multicolumn{1}{|c|}{}                               & \tiny{max} & {\scriptsize{2.76}}                      & {\scriptsize{1.28}}  & \multicolumn{1}{r|}{\scriptsize{1,144.46}} & \scriptsize{0.87}                     \\ \cline{2-6} 
                    \multicolumn{1}{|c|}{}                               & \tiny{avg} & {\scriptsize{3.27}}                   &{\scriptsize{1.72}} & \multicolumn{1}{r|}{\scriptsize{885.66}}   & \scriptsize{1.13}                     \\ \hline
                \end{tabular}
                \caption{Total inference time in milliseconds and FPS, as the sum of pre-processing, inference, and post-processing times, for both single object and multi objects scenarios with the corresponding \ac{FPS} value. 
                }
                \label{tab:times}
            \end{table}

            In general, since we propose a new dataset it is hard to compare the resulting values to other \acs{6D} pose datasets.

        \subsubsection{Benchmark with Literature\label{sssec:benchmark} to fairly compare with \ac{YOLO}-6D and EfficientPose, we train and test all the methods using simulated underwater data (see Section~\ref{sec:data_collection}). We select a subset of $250$ images, from the $500$ extracted in Section~\ref{sssec:baseline}, corresponding to various synthetic scenes of hotstab (symmetric object) and the jug (asymmetric object).} 

            \textbf{Training:} 
            We iterated on different combinations of batch sizes, learning rates and Adam momentum as optimiser for the training of \ac{YOLO}-6D and EfficientPose. The performance reported in this section for \ac{YOLO}-6D is achieved with batch size equal to $32$, an adaptive learning rate (it starts with $0.0001$ for $10{,}000$ epochs. For EfficientPose, the best performance is achieved with a batch size equal to $1$, a learning rate of $0.0001$ and $500$ epochs. For our proposal, we train as detailed in Section~\ref{ss:AAE}. However, to ensure comparability with the other two methods, we opted for lighter versions by selecting $\phi$ equal to 0 as scaling hyperparameter. 

            \textbf{Results:} 
            Table~\ref{tab:method_comparison} shows the results of comparing \ac{YOLO}-6D, EfficientPose and our method using the \ac{BOP} metrics. In the case of EfficientPose, contrary to the high performance on the Linemod Dataset, it achieves poor results in our dataset. We observe that EfficientPose's 2D detection achieves only $77\%$, consequently compromising the 6D pose estimation. \cite{govi2023uncovering} presents an exhaustive study on the potential causes for object detection failure in EfficientPose.
           
            \begin{table}[!t]
                \centering
                \begin{tabular}{|c||c|c|c|}
                %{ |p{3cm}||p{2.5cm}|p{2.5cm}|p{1.5cm}|  }
                 \hline  
                   & ADD & ADI & mAP \\
                  \hline
                %     \begin{tabular}{c|c|c|c}
                    %      & \scriptsize{Test on the same dataset} & \scriptsize{Test on the other dataset} & Average \\
                    % \hline
                    \multicolumn{4}{|c|}{Symmetric Object (hotstab)}\\
                    \hline
                     \scriptsize{EfficientPose}& 1.32\%  & 8.65\% & 75.21\% \\
                     %\hline
                     %Yolo-6D  & 3.79\% & 19.22\% & - \\
                     \hline
                     \scriptsize{YOLO-6D} & 9.58\%  & 36.71\% & 77\%\\
                     \hline
                     \scriptsize{\textbf{YOLOv4+AAE}} & 11.4\% & \textbf{44.0\%} & \textbf{99\%} \\
                     \hline
                     \multicolumn{4}{|c|}{Asymmetric Object (jug)}\\
                     \hline
                     \scriptsize{EfficientPose}& 23\%  & 54.8\% & 73.41\% \\
                     %\hline
                     %Yolo-6D  & 3.79\% & 19.22\% & - \\
                     \hline
                     \scriptsize{YOLO-6D} & 26.50\%  & 58.44\% & 81.6\%\\
                     \hline
                     \scriptsize{\textbf{YOLOv4+AAE}} & \textbf{29.2\%} & \textbf{78.2\%} & \textbf{99.5\%}\\
                     \hline
                \end{tabular}
                \caption{EfficientPose, \ac{YOLO}-6D, and \ac{YOLO}v4+\ac{AAE} comparison using \ac{BOP} metrics for the hotstab and the jug objects. The \ac{ADD} and \ac{ADI} recall is computed with a threshold $k_m = 0.1$. The mAP is computed with a $0.5$ of \ac{IOU}.}
                \label{tab:method_comparison}         
            \end{table}

            \subsubsection{Real data error pose estimation\label{ss:error_estimation}}
            to test the robustness of our method's performance in real scenes, despite being trained in simulation, we calculate the error pose using the ground truth data (see Section~\ref{ss:bundle_gt}). Moreover, this error pose estimation offers a baseline for evaluating future 6D pose estimation methods that require real ground truth labelled data for training.
            From the $87{,}100$ collected images in our dataset, $83{,}701$ contained the bundle described in Section~\ref{ss:bundle_gt}. From this subset, we selected $20\%$ of the data to process the translation and rotation error in Table~\ref{tb:transAndRotError}. Specifically, we calculate the Euclidean distance between the position and orientation estimation error of our proposed pipeline in Fig.~\ref{fig:diagram} with respect to the ground truth pose extracted from the bundle holder. The evaluation subset is available in our website\footnote{Our dataset is available at \url{https://bit.ly/3LZYvyJ}\label{ft:dataset}}.
            
            As seen by the error estimation, the position error median is approximately $20mm$ while the normalised orientation error (between $0$\textdegree~and $360$\textdegree) is around $30$\textdegree~for the different objects. It is worth noting that higher error values correspond to the rotation error rather than the translation one. Particularly, this difference is more evident for the cube and mug objects. This result can be attributed to the texture-less and symmetrical nature of objects. While the pose estimation has room for improvement, particularly for the object rotation, the values are consistent with the original \ac{AAE} proposal in \cite{AAE5}, demonstrating the method's consistency across symmetric objects.
            
            %------------------------------
            % \begin{figure}[b!]
            %     \centering
            %     \includegraphics[width=9cm]{Figures/bundle_ground_truth.png}
            %     \caption{Median and variance data for pose error representation of our pipeline using the pose extracted from AprilTag bundles as ground truth for each of our settings. Translation in purple in mm and rotation in green in degrees.
            %     \label{fig:transAndRotError}}
            % \end{figure}
                %----------------------

            %-------------------------
            \begin{table}[ht]
                \fontsize{6.5}{6.5}\selectfont
                \begin{center}
                    \begin{tabular}{|c|c|c|c|c|c|c|c|c|c|}
                    \hline
                    \scriptsize{\textbf{Scenes}} &
                    \multicolumn{2}{c|}{\scriptsize{\textbf{Hotstab}}} & \multicolumn{2}{c|}{\scriptsize{\textbf{Cube}}} & \multicolumn{2}{c|}{\scriptsize{\textbf{Mug}}} & \multicolumn{2}{c|}{\scriptsize{\textbf{Jar}}} \\
                    \hline
                    Asphalt & $t_e$ & $r_e$ & $t_e$ & $r_e$  & $t_e$ & $r_e$  & $t_e$ & $r_e$ \\
                    \hline
                    Dry shadow & 22.3 & 25 & 23.7 & 23.5 & 21.1 & 25 & 23.3 & 21 \\
                    \hline
                    Dry dirt & 16 & 23 & 28.2 & 21.2 & 15 & 24.3 & 20.1 & 22.4 \\
                    \hline
                    UW partly occluded & 22.1 & 29 & 24.2 & 29 & 21.1 & 24 & 19 & 24 \\
                    \hline
                    UW with LED & 21 & 20.9 & 24 & 30 & 19 & 25 & 20 & 24 \\
                    \hline
                    Grass & 20 & 22 & 29.8 & 31.4 & 20 & 27 & 21 & 21.3 \\
                    \hline
                    Dry white & 20 & 27 & 25 & 32.1 & 20 & 28.1 & 20 & 27.1 \\
                    \hline
                    Dry blue & 15 & 25 & 23.6 & 29 & 22 & 21.8 & 24.2 & 23 \\
                    \hline
                    UW with shadow & 17 & 26.3 & 18 & 25 & 19 & 24.3 & 22.7 & 23 \\
                    \hline
                    UW with plants & 20 & 20.3 & 24 & 25.1 & 20 & 23 & 19.9 & 20 \\
                    \hline
                \end{tabular}                
                \end{center}
                \caption{Median data for pose error representation of our pipeline using the pose extracted from AprilTag bundles as ground truth for each of our settings. Translation error, $t_e$ in mm and rotation error, $r_e$ in degrees. \label{tb:transAndRotError}}
            \end{table}
            %-------------------------

    \subsection{Underwater Manipulation Use Case}\label{ss:evaluation}

        % robotic manipulation
        % maybe mapping for objects locations when exploring in an environment (under consideration)
        
        In order to test the robustness of our object pose estimation pipeline we tested it with our underwater manipulation setup. Our testing setup consisted of a static robotic arm, Reach Bravo with 7~\ac{DoF} on a bench structure that served as the skeleton to hold the arm inside our $2m$ depth pool. We placed a surface with a solid background and attached a weight to our target object to avoid pose oscillations due to the object's buoyancy. We attached a realsense $d455$ camera on the arm's skeleton to have a complete view of the arm and the object. From the camera we extract solely the \ac{RGB} data as input to our pipeline described in Sections ~\ref{ss:yolov4} and \ref{ss:AAE}.
        
        % Once we have obtained the \acs{6D} pose from the pipeline proposed in this letter, we use our in-house developed assistant for manipulation tasks to extract a reaching goal and a motion plan as well as to concatenate the task. Namely, our manipulation interface takes high-level commands from a human to author and manage the manipulation task. In the example of reaching for an object, the manipulation assistant takes the semantic label of the detected object and allows the user to select a series of tasks. For the pipeline proof-of-concept, we decided to use a single object in the scene to ease the manipulation goal decision-making process. 

\begin{figure}[tb!]
            \centering
            \includegraphics[width=8cm]{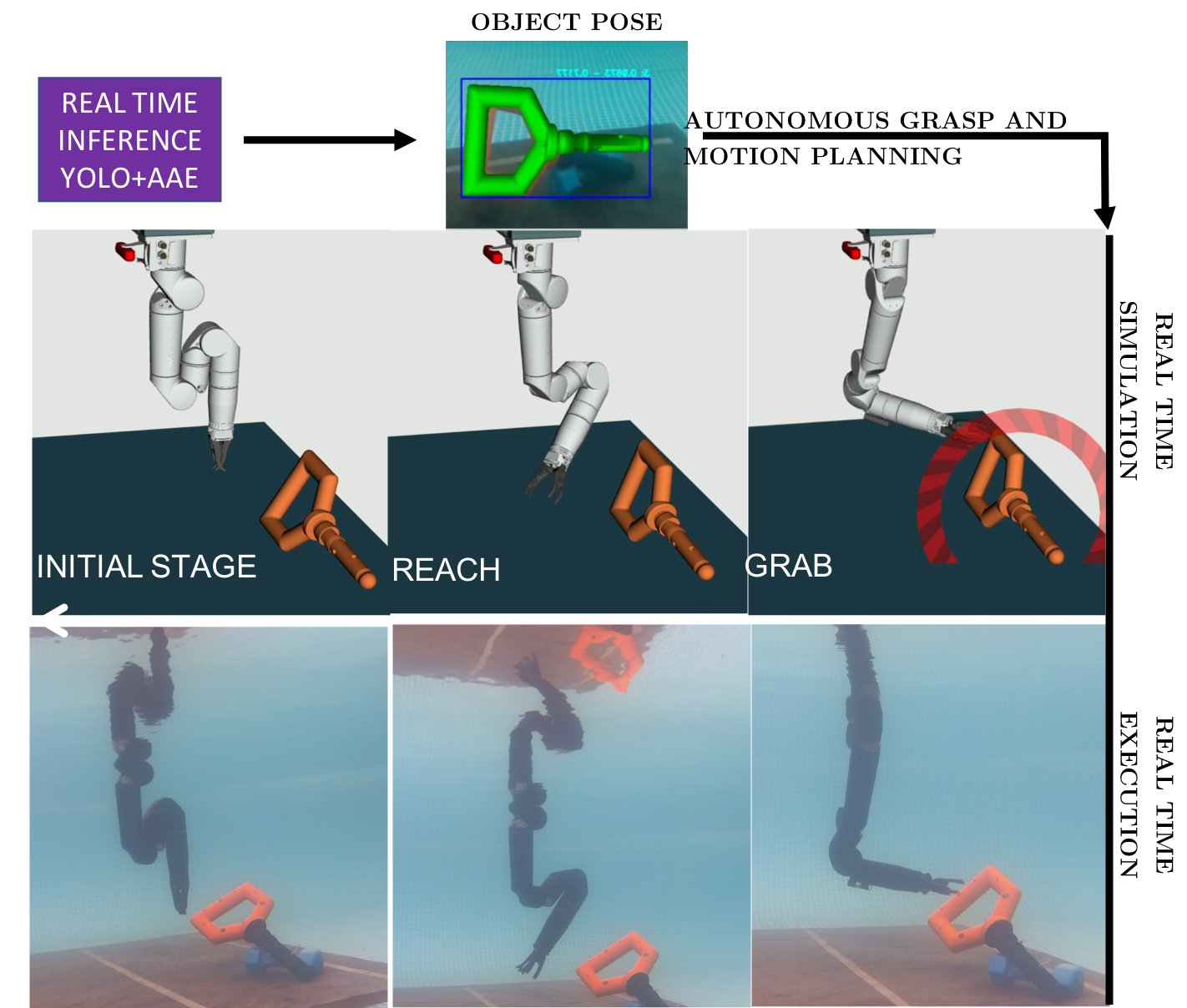}
            \caption{Our proposed pipeline in an underwater manipulation example. Given the object pose of an object, we autonomously generate (i)~grasp configurations on the objects and (ii)~a motion plan to reach and grab the target object.
            \label{fig:manipulation_uc}}
        \end{figure}

        Once we have obtained the \acs{6D} pose from the pipeline proposed in this letter, we use our in-house developed assistant for manipulation tasks to aid in taking authoring high-level manipulation commands. The assistant then extracts and concatenates in a single task all required reaching goals and motion plans. In the example of reaching for an object, the manipulation assistant takes the semantic label of the detected object and allows the user to select a series of tasks. For the pipeline proof-of-concept, we decided to use a single object in the scene to ease the manipulation goal decision-making process. Using our user interface, we select \texttt{Open Gripper} and \texttt{Reach Hotstab} (see illustration in Fig.~\ref{fig:manipulation_uc}). Out of $10$ different poses, inside the arm's working space, $6$ resulted in accurate poses for the arm to successfully reach\footnote{Experiments: \url{https://www.youtube.com/watch?v=xPAbxwh5JGM}}. The failed trials are attributed to flickering in the \ac{AAE} pose estimation. These oscillations are the result of unavoidable reflections from the water surface due to the shallowness of our testing environment. Nonetheless, for future improvements, a pose tracker and filtering could be put in place to ensure grabbing the object. Moreover, in some of the frames, we noticed that the \ac{2D} object detection model would detect some outlier objects in the scene. To ensure the manipulator did not plan trajectories to false positive poses, we selected a priori the target object. However, the 2D object detection model could be further improved by including background data with similar geometries that do not belong to our objects of interest.
        
%% file: Sections/4_conclusions.tex
\section{Conclusions and Future Work}\label{sec:conclusions}
    We propose a method for underwater pose estimation using  \ac{RGB} data that copes with diverse environments. We summarise our contribution as twofold: (i)~a publicly available dataset consisting of $4$ objects in $10$ different real scenes and annotations for object detection and pose estimation, as well as, (ii)~an object pose estimation pipeline that adapts to different light conditions and heterogeneous backgrounds in real scenes while being trained in simulated data.

    Our dataset consists of some challenging objects with a symmetrical shape and poor texture. Regardless of such object characteristics, our proposed method outperforms alternative model based \ac{6D} pose estimation methods. Our results suggest potential for generalisation with symmetrical, asymmetrical and texture-less object representations. We attribute this performance to the robustness achieved in the \ac{AAE} through our hyperparametrisation stage. We successfully used our proposed pipeline for a reaching task using an underwater manipulator demonstrating its robustness. It is also worth mentioning that our pipeline has some inherent limitations given the adopted self-supervised learning approach. One of these limitations is the scalability of the pipeline as target objects in the scene increase, since there should exist a \ac{AAE} model per object. However, we envision using our pipeline for manipulation purposes in which there are limited target objects at a close-reaching range.
    
    This work opens multiple avenues for further research, such as facilitating the object generalisation through an online \ac{3D} shape estimation algorithm, and stabilising the pose detection with a pose tracker and filtering to avoid oscillations in the pose estimation. We enable such future work by making our proposed pipeline and dataset publicly available.
    
% \section{Acknowledgements}

%     Thanks to SpinItalia\footnote{\url{http://www.spinitalia.com/}} for facilitating testing locations and hardware for experiments in Italy.